\documentclass[10pt,twocolumn,letterpaper]{article}

\usepackage{icb}
\usepackage{times}
\usepackage{epsfig}
\usepackage{graphicx}
\usepackage{amsmath}
\usepackage{amssymb}
\usepackage{comment}
\usepackage[flushmargin]{footmisc}

\icbfinalcopy 

\ificbfinal\fi

\makeatletter 
\def\ps@IEEEtitlepagestyle{ 
\def\@oddfoot{\mycopyrightnotice} 
\def\@evenfoot{} 
} 
\def\mycopyrightnotice{ 
{\hfill \footnotesize 978-1-7281-3640-0/19/\$31.00 \copyright 2019 IEEE\hfill} 
} 
\makeatother 

\begin{document}

\title{Universal Material Translator: Towards Spoof Fingerprint Generalization}

\author{Rohit Gajawada\thanks{These authors have contributed equally.} $^\dagger$, Additya Popli\footnotemark[1] $^\dagger$, Tarang Chugh$^\ddagger$, Anoop Namboodiri$^\dagger$ and Anil K. Jain$^\ddagger$\\
$^\dagger$IIIT Hyderabad, India \\
$^\ddagger$Michigan State University, USA\\
{\tt\small  \{rohit.gajawada@students.iiit.ac.in, additya.popli@research.iiit.ac.in,}\\
{\tt\small chughtar@cse.msu.edu, anoop@iiit.ac.in, jain@cse.msu.edu\}}
}

\maketitle
\thispagestyle{empty}


\begin{abstract}
    Spoof detectors are classifiers that are trained to distinguish spoof fingerprints from bonafide ones.
    However, state of the art spoof detectors do not generalize well on unseen spoof materials. This study proposes a style transfer based augmentation wrapper that can be used on any existing spoof detector and can dynamically improve the robustness of the spoof detection system on spoof materials for which we have very low data. Our method is an approach for synthesizing new spoof images from a few spoof examples that transfers the style or material properties of the spoof examples to the content of bonafide fingerprints to generate a larger number of examples to train the classifier on. We demonstrate the effectiveness of our approach on materials in the publicly available LivDet 2015 dataset and show that the proposed approach leads to robustness to fingerprint spoofs of the target material.
\end{abstract}

{\let\thefootnote\relax\footnotetext{\mycopyrightnotice}}

\section{Introduction}
Fingerprints have been widely used for identification of individuals for more than a century. Their uniqueness and persistence makes it one of the most reliable forms of biometric traits that have been used to augment security in a variety of applications. However, with the widespread use of fingerprint recognition systems, they are now a prime target of attackers. Fingerprint spoofs (i.e. fake/gummy fingers) can be fabricated using commonly available materials such as Play Doh, Gelatin etc. with high fidelity, making them hard to distinguish from bonafide fingerprints. These materials can easily fool the existing spoof detectors especially if they were not included in the training. With a plethora of available materials that could be used to fabricate spoofs, it is important to be able to learn the concept of what a specific spoof looks like from just a few samples of it.

\begin{figure}[t]
\centering
    \includegraphics[scale=0.46]{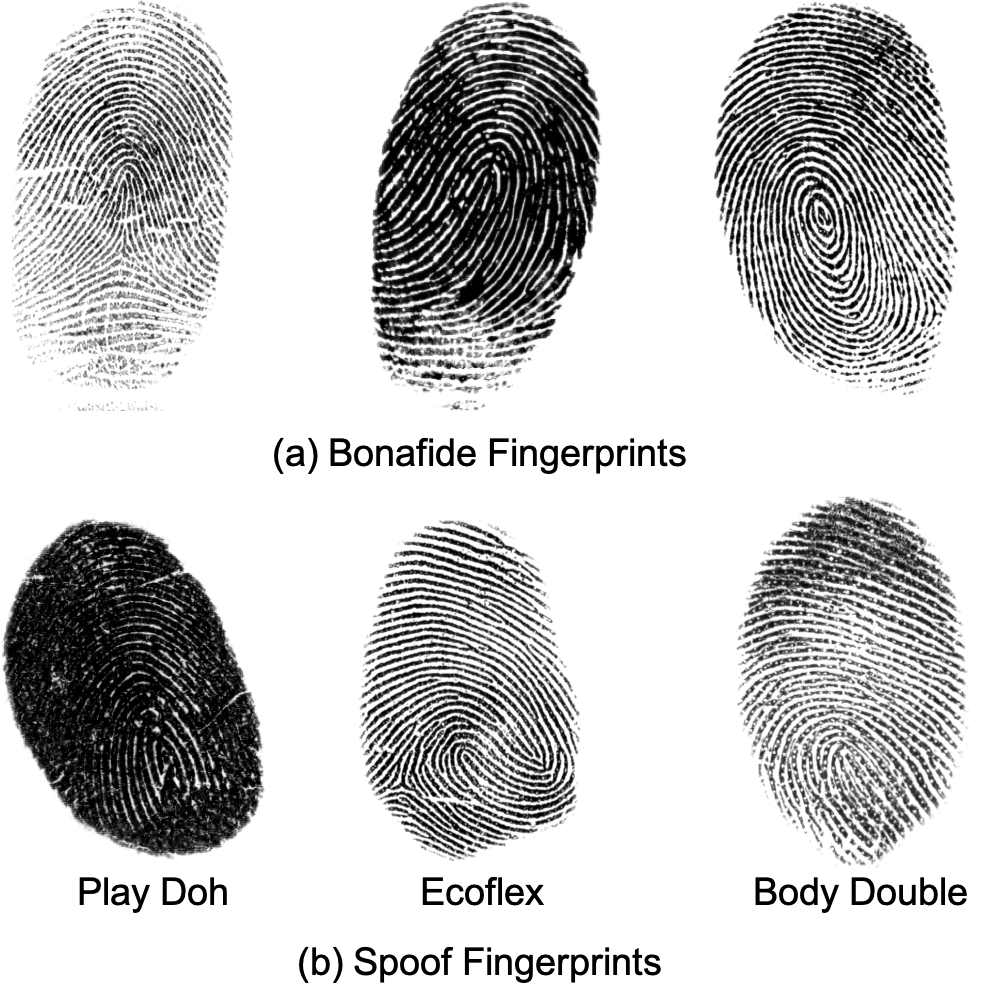}
\caption{(a) Bonafide fingerprint samples and (b) Acquired spoof fingerprint samples fabricated using different materials.}
\vspace{-0.5cm}
\label{fig:intro}
\end{figure}

\begin{figure*}[t!]
\centering
\includegraphics[width=\linewidth]{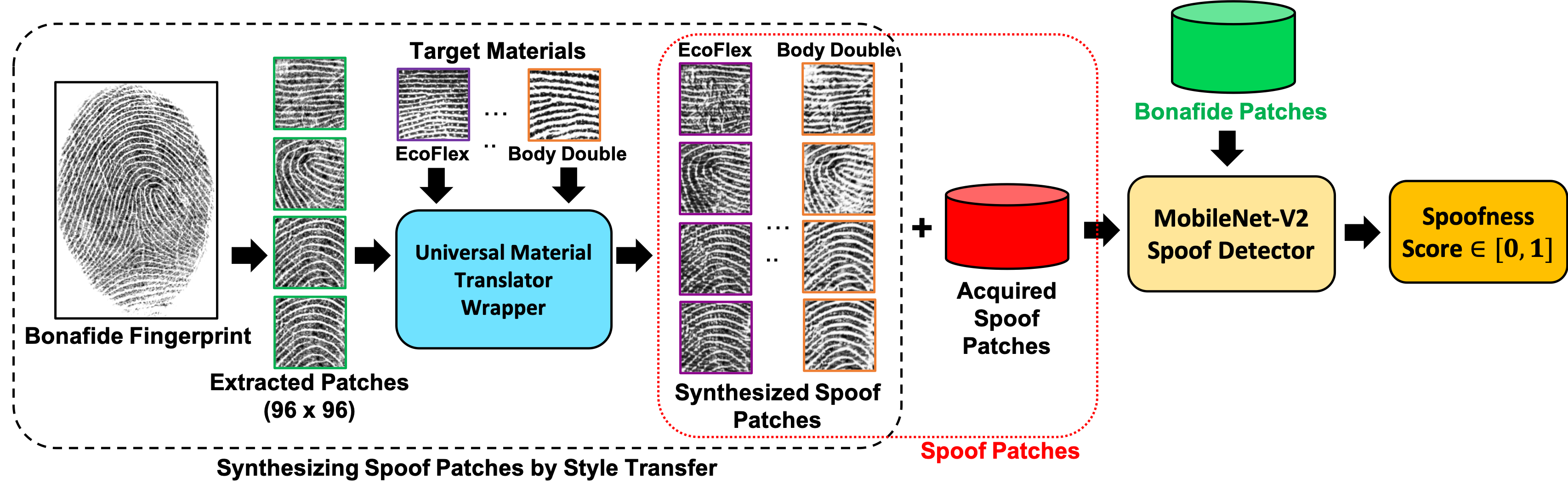}
\caption{An overview of the proposed approach utilizing Universal Material Transfer (UMT) wrapper for synthesizing spoof patches.}
\label{fig:overview}
\vspace{-3mm}
\end{figure*}

One of the major limitations of many published anti-spoofing methods is their poor generalization performance across novel spoof materials, that were not seen during training. To generalize an algorithm's effectiveness across spoof fabrication materials, called \textit{cross-material} performance, some studies have approached spoof detection as an \textit{open-set problem}\footnote{Open-set problems address the possibility of spoof classes during testing, that were not seen during training. Closed-set problems, on the other hand, evaluate only those spoof classes that the system was trained on.}. ~\cite{rattani2015open} applied the Weibull-calibrated SVM (W-SVM) to detect spoofs made of new materials. ~\cite{ding2016ensemble} trained an ensemble of multiple one-class SVMs using textural features extracted from only bonafide fingerprint images.

Although, CNN-based approaches outperform the earlier mentioned spoof detection approaches utilizing hand-crafted features on publicly available datasets~\cite{ghiani2017review}, these require large amount of training data to avoid over-fitting. With the advent of new fabrication techniques and novel materials, it is increasingly challenging to include all spoof fabrication materials and generate large datasets for each new material for training. In this study, we show that as few as 5 sample spoof fingerprint images can go a long way to improve the robustness of spoof detectors towards a novel fabrication material. It enables a framework, called Universal Material Translator (UMT) that can be used for cross-material spoof fingerprint style transfer. Moreover, we show that these style transferred images can be used effectively for augmenting CNN-based spoof detectors, significantly improving their performance against a novel material, while retaining its performance on known materials.

\section{Related Work}
Fingerprint spoof detection has been approached from both hardware-based and software-based solutions. Hardware solutions usually rely upon detecting physical liveness characteristics of a human finger, such as blood flow ~\cite{lapsley1998anti}, human odor~\cite{baldisserra2006fake} and multi-perspective images \cite{tpamijosh}. Software approaches do not require specialized sensors as they work with features extracted from captured fingerprint images to perform spoof detection.

Conventional software-based solutions~\cite{marcialis2010analysis, marasco2012combining, ghiani2012fingerprint, ghiani2013fingerprint} extract hand-crafted features based on texture, physiology or anatomy of the input fingerprints to detect spoofs. 
Recently, learning-based approaches (CNNs in particular) have been utilized resulting in a higher degree of accuracy than previous works. Nogueira et al.~\cite{nogueira2016fingerprint} uses transfer learning from object-recognition CNNs pre-trained on ImageNet \cite{imagenet}, and fine-tuned with fingerprints for spoof detection. Pala et al.~\cite{pala2017deep} uses randomly selected patches to train a CNN architecture with triplet loss. Chugh et al.~\cite{chugh2018fingerprint} utilizes local patches extracted and aligned using fingerprint minutiae to train a two-class CNN to detect spoof fingerprints. It is based on the premise that noise related to spoof fabrication introduces artifacts, that is pronounced around anomalous points in a fingerprint image i.e., \textit{minutiae points}. It has performed better with novel spoof materials than previous approaches noted above.

Realistic image synthesis is a difficult problem. Early non-parametric methods \cite{patchmatch, sketch2photo, efros_texture} find difficulty in generating images with textures that are unseen during training. Machine learning has been very effective in this regard both in terms of realism and generality. \cite{neuralstyle} and \cite{gatyssynth} perform artistic style transfer, combining the content of an image with the style of any other by minimizing the feature reconstruction loss and a style reconstruction loss which are based on features extracted from a pretrained convolutional network at the same time. Their method produces high quality results, but is computationally expensive since each step of the optimization problem requires a forward and backward pass through the pretrained
network. \cite{johnsonstyle, wangstyle, ulyanovstyle, listyle} train a feed-forward network to quickly approximate solutions to this optimization problem. \cite{improvedtextnetworks, condin, adainmain} use methods based on feature statistics to perform style transfer.
\cite{elgammal2017can} applied GANs to generate artistic images. ~\cite{isola2016image} used conditional adversarial networks to learn the loss for image to image translation.
 \cite{texturegan} learnt to synthesize objects consistent with texture suggestions.
 Our approach utilizes the approach in \cite{adainmain} for designing the proposed universal material translator that is capable of producing realistic fingerprint patches containing friction ridge information (content) of bonafide patches and style information from target material spoof patches.

\begin{figure*}[t!]
\vspace{-0.6cm}
\centering
\includegraphics[width=\linewidth]{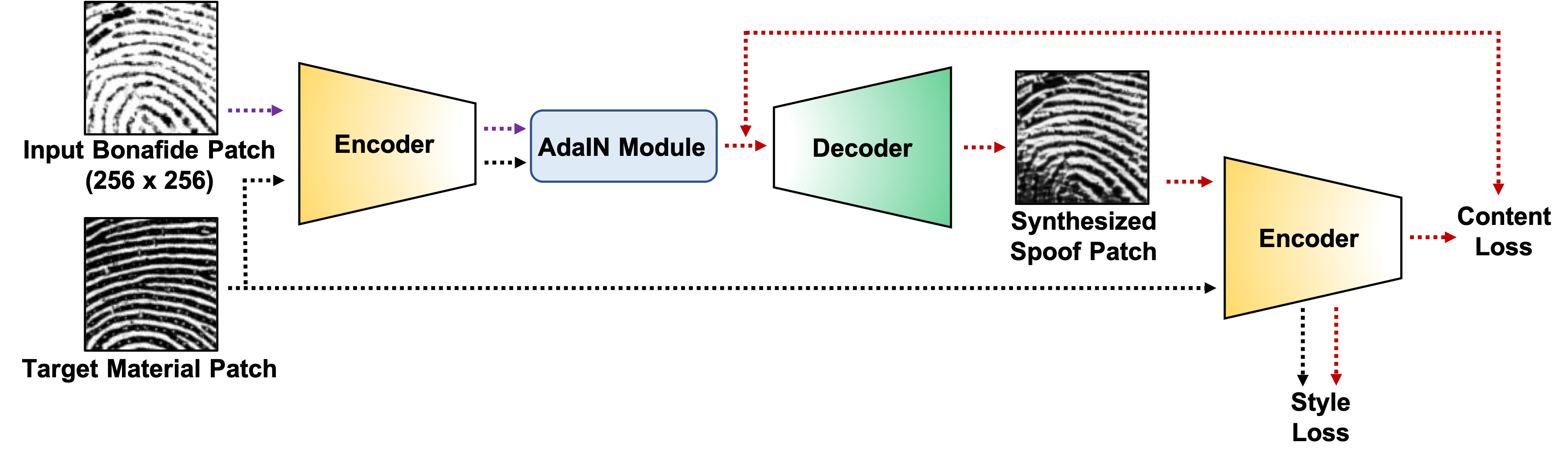}
\caption{Training pipeline of the Universal Material Translator (UMT) wrapper. An AdaIN module is used for performing the style transfer in the encoded feature space. The same VGG-19 encoder is used for computing content loss and style loss.}
\label{fig:archgan}
\end{figure*}


\section{Universal Material Translator} 
Studies in~\cite{nogueira2016fingerprint, marasco2011robustness, rattani2015open} have shown that when a spoof detector is evaluated on spoofs fabricated using materials that were not seen during training, there can be up to a three-fold increase in the spoof detection error rates. \cite{chugh2019fingerprint} identified a subset of spoof fabrication materials based on their optical and mechanical properties that are crucial to train a robust spoof detector. However, with the advent of new fabrication techniques and novel materials, it is prohibitively expensive and time-consuming to generate large-scale datasets for each new material for training CNN-based approaches. Therefore, it is crucial to generate high quality synthetic fingerprint spoof images to address the lack of spoof training data.

In order to move towards this goal, we desire to generate synthetic spoof images that are similar in material style to the spoofs for which we only have very limited number of samples. It is possible that we might have only a few spoof images of a particular material due to expensive fabrication process or use of rare materials, and we cannot simply use these few images to train a deep CNN-based spoof classifier. 

Towards this goal, we propose {\bf Universal Material Translator (UMT)} wrapper, a setup that extracts the material style characteristics from a few acquired spoof fingerprint images\footnote{\textit{Acquired} spoof fingerprint images refer to the fingerprint impressions of physical spoof specimens that are captured using a fingerprint reader, as opposed to \textit{synthesized} spoof fingerprint images which are generated images using the proposed approach.} and transfers it to the available large-scale bonafide image database to generate synthetic yet realistic spoof images in the desired target material style while retaining the friction ridge information of the source fingerprint. 
In our experiments, we utilize local patches of size $96 \times 96$ for style transfer instead of whole images. Figure~\ref{fig:overview} presents an overview of the proposed approach utilizing Universal Material Translator wrapper for synthesizing spoof patches in target material style. From here onwards, we refer spoofs acquired using fingerprint readers as spoofs and spoofs fabricated using the proposed approach as synthesized spoofs.

\subsection{Material Style Transfer}
\cite{improvedtextnetworks} used an InstanceNorm layer to normalize feature statistics across spatial dimensions given a style image, per channel per sample. The equation for InstanceNorm is given below: 
\begin{equation} \label{eqn:basic} IN(x) = \alpha\Big(\dfrac{x - \mu(x)}{(\sigma(x))}\Big) + \beta  \end{equation} 
In Equation 1, $x$ is the input feature space, $\mu(x)$ and $\sigma(x)$ are the mean and variance parameters of that feature space, respectively. It was observed that changing the affine parameters $\alpha$ and $\beta$ (while keeping convolutional parameters fixed) leads to variations in the style of the image, and the affine parameters could be learned for each particular style. This motivated \cite{condin}, which learns alpha and beta values for each feature space and style pair. However, this required retraining of the network for each new style. \cite{adainmain} replaced the InstanceNorm layer with an AdaIN layer, which can directly compute affine parameters from the style image, instead of learning them -- effectively transferring style by imparting second-order statistics from the style image to the content image, through the affine parameters. We follow the same approach as described in \cite{adainmain} in our paper, by transferring feature statistics from the target material spoof to the input content bonafide in the feature space. As described in AdaIN, we apply instance normalization on the input content bonafide feature space however not with learnable affine parameters. The channel-wise mean and variance of the content bonafide's feature space is aligned to match those of the target material spoof's feature space. This is done by computing the affine parameters from the target material spoof feature space in the following manner:
\begin{equation} AdaIN(x,y) = \sigma(y)\Big(\dfrac{x - \mu(x)}{(\sigma(x))}\Big) + \mu(y)  \end{equation}

\noindent In Equation 2, the content bonafide patch's feature space is $x$ and the target material spoof patch's feature space is $y$. In this manner, $x$ is normalized with $\sigma(y)$ and shifted by $\mu(y)$. Our synthetic spoof generator $G$ is composed of an encoder $Enc$ and a decoder $Dec$. For the encoder, we use the first few layers of a pre-trained VGG network similar to~\cite{johnson2016perceptual}. The weights of this network are frozen throughout all stages of the setup. Given the content bonafide patch as $c$ and the target material spoof patch as $m$, then $x$ is $Enc(c)$ and $y$ is $Enc(m)$. The desired feature space is obtained as: \begin{equation} t = AdaIN(Enc(c),Enc(m)) \end{equation}

\noindent We use $Dec$ to take $t$ as input to produce $Dec(t)$ which is the final synthesized spoof stylized in the desired material.

In order to ensure that our synthesized spoof patches i.e $Dec(t)$ are matching the style statistics of the target material spoof, we apply a loss $L_s$ similar to~\cite{johnson2016perceptual} given as:
\begin{equation} \begin{array} { r } { \mathcal { L } _ { s } = \sum _ { i = 1 } ^ { L } \left\| \mu \left( \phi _ { i } ( g ( t ) ) \right) - \mu \left( \phi _ { i } ( s ) \right) \right\| _ { 2 } + }\\ { \sum _ { i = 1 } ^ { L } \left\| \sigma \left( \phi _ { i } ( g ( t ) ) \right) - \sigma \left( \phi _ { i } ( s ) \right) \right\| _ { 2 } } \end{array} \end{equation}


\noindent where each $\phi_i$ denotes a layer in the VGG-19 network we use as encoder. We pass Dec(t) and m through Enc and extract the outputs of $\it relu1\_1$, $\it relu2\_1$, $\it relu3\_1$ and $\it relu4\_1$ layers for computing $L_s$.
 To ensure that the synthesized spoof patches retain fingerprint (friction ridge) content from bonafide patches, we use a content loss $L_c$ which is computed as the euclidean distance between the features of the synthesized spoof patches \textit{i.e.} $Enc(Dec(t))$ and the features of the bonafide patches with the target style ($t$).


\noindent The effective style loss is given as: 
\begin{equation}
    L = \lambda_c L_c + \lambda_s L_s
\end{equation}

\cite{demystneural} goes into the theory behind why neural style transfer works so well, by interpreting it as a domain adaptation problem. Although we do not follow the gram-matching approach, our style transfer still involves second-order feature statistics transfer, similar to what is achieved by gram matrix matching.

\begin{figure}[t]
\centering
\includegraphics[width=75mm]{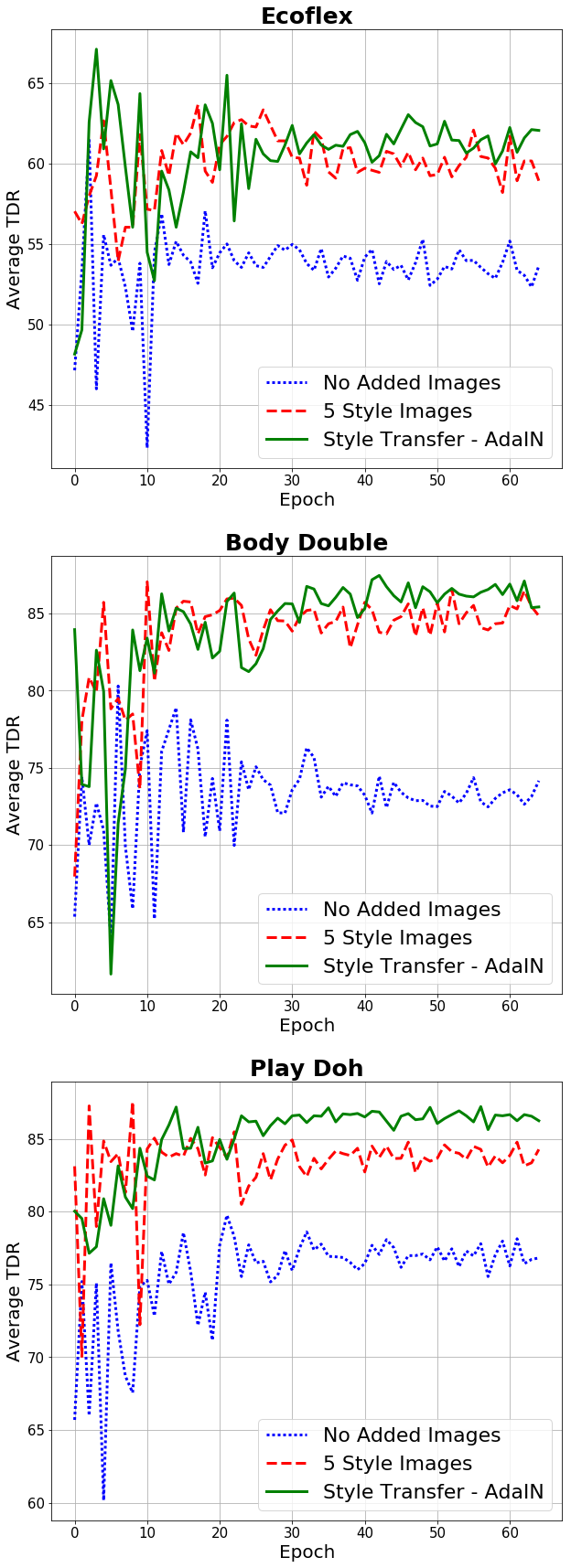}
\caption{TDR\% on unknown materials at FDR = $0.1\%$}
\label{fig:introdiag}
\vspace{-0.5cm}
\end{figure}


\begin{figure*}
\vspace{-0.6cm}
\centering
\includegraphics[width=\linewidth]{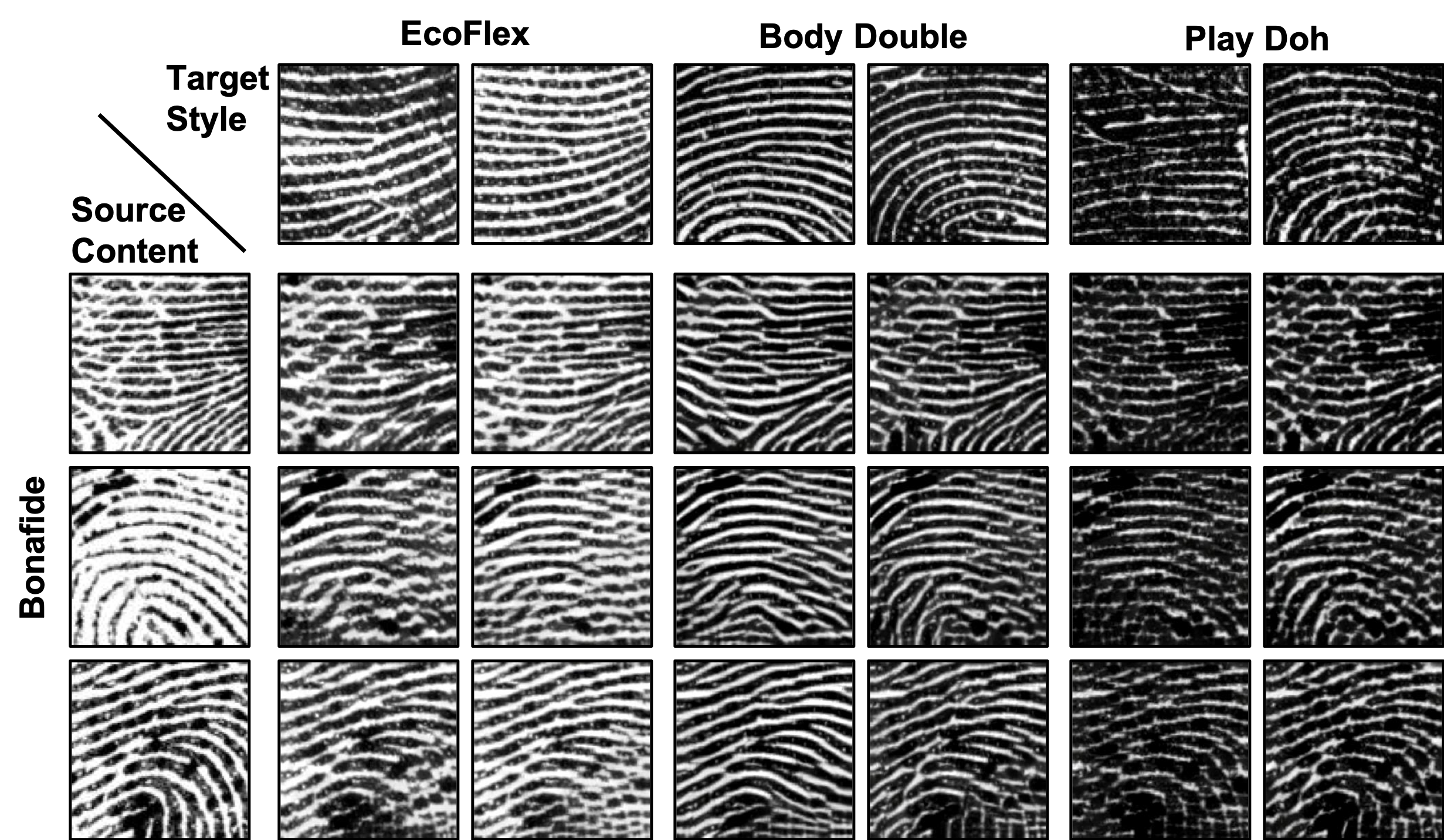}
\vspace{-0.2cm}
\caption{Spoof patches synthesized from bonafide patches (first column) by our approach conditioned on a target material (first row).}
\label{fig:spoofgan}
\end{figure*}

\section{UMT-Wrapper for Spoof Robustness}
Once we obtain a spoof synthesizer $G$, we use the existing dataset of bonafide patches and generate a dataset of synthesized spoof patches conditioned on the material $M$. In our experiments, the bonafide fingerprints used for generating new spoofs in the target material style are the materials that we use in our training set in that particular experiment.
We combine this synthetic data with the existing training data of LivDet15~\cite{mura2015livdet} for training our classifier. Note when evaluating the performance on an unseen material, spoofs of that particular material are removed from the training dataset of the LivDet15~\cite{mura2015livdet} split.
Our approach acts like a wrapper on top of any existing spoof detector. If our approach is applied to a spoof detector, it can successfully make it more robust to spoofs of particular target materials just seen during training.

For our classification experiments, we use the Mobilenet V2 \cite{mobilenetv2} pretrained on ImageNet \cite{imagenet} as it is faster than other architectures like Inception V3 \cite{incv3} used in published studies. We follow a training procedure similar to \cite{chugh2018fingerprint}.

\section{Experiments and Results}
\subsection{Datasets}
We use the Crossmatch sensor of the LivDet15 \cite{mura2015livdet} dataset in our experiments. For each image in the training and testing dataset, we perform Otsu's Global Thresholding \cite{otsu} followed by a set of morphological operations to obtain a segmentation mask for the fingerprint similar to how it is done in \cite{MinNet}. To further remove any background noise, we find the largest connected component in the foreground part of this resultant image. We generate 30 patches of size $150 \times 150$ which lie within the foreground of this resultant image. For all the patches, we compute the orientation similar to how it is done in \cite{MinNet} with a window of $64$ and stride of $32$. We perform alignment on each patch by taking the trimmed mean of the orientations, rotating the patch by this mean orientation and doing a center crop of $96 \times 96$ from the resultant image.

\begin{table*}[t]
\centering
\caption{Cross-material performance (TDR (\%) @ FDR = $0.1\%$) using EcoFlex, Body Double, and Play Doh spoof materials, without and with using synthesized spoof patches in training.}
\label{tab:crossresults}
\resizebox{\linewidth}{!}{
\begin{tabular}{ | p{12.3cm} | p{3.8cm} | p{2cm} | } \hline 
\textbf{Training Set} & \textbf{Testing Set} & \textbf{ TDR (\%) @ FDR = 0.1\%}  \\ \hline \hline
Bonafide vs. [Play Doh + Body Double] & Bonafide vs. EcoFlex & $53.56$ $\pm$ $3.05$ \\ \hline
Bonafide vs. [Play Doh + Body Double + 150 Spoof EcoFlex Patches*] & Bonafide vs. EcoFlex & $59.99$ $\pm$ $6.23$ \\ \hline
Bonafide vs. [Play Doh + Body Double + 150 Spoof EcoFlex Patches + 15K Synthesized EcoFlex Patches] & Bonafide vs. EcoFlex & $\textbf{61.40}$ $\pm$ $\textbf{3.73}$ \\ \hline \hline

Bonafide vs. [Play Doh + EcoFlex] & Bonafide vs. Body Double & $73.20$ $\pm$ $8.59$ \\ \hline
Bonafide vs. [Play Doh + EcoFlex + 150 Spoof Body Double Patches*] & Bonafide vs. Body Double & $85.01$ $\pm$ $1.96$ \\ \hline
Bonafide vs. [Play Doh + EcoFlex + 150 Spoof Body Double Patches + 15K Synthesized Body Double Patches] & Bonafide vs. Body Double & $\textbf{86.24}$ $\pm$ $\textbf{2.28}$ \\ \hline \hline

Bonafide vs. [EcoFlex + Body Double] & Bonafide vs. Play Doh & $76.99$ $\pm$ $4.08$ \\ \hline
Bonafide vs. [EcoFlex + Body Double + 150 Spoof Play Doh Patches*] & Bonafide vs. Play Doh & $83.90$ $\pm$ $4.64$ \\ \hline
Bonafide vs. [EcoFlex + Body Double + 150 Spoof Play Doh Patches + 15K Synthesized Play Doh Patches] & Bonafide vs. Play Doh & $\textbf{86.49}$ $\pm$ $\textbf{1.56}$ \\ \hline
\end{tabular}
}
\flushleft{\footnotesize{$^*$150 Spoof EcoFlex / Body Double / Play Doh Patches are generated from only 5 Spoof EcoFlex / Body Double / Play Doh images, respectively.}}
\end{table*}

\begin{table*}[t]
\centering
\caption{Known-material performance (TDR (\%) @ FDR = $0.1\%$) using EcoFlex, Body Double, and Play Doh spoof materials, without and with using synthesized spoof patches in training.}
\label{tab:knownresults}
\resizebox{\linewidth}{!}{
\begin{tabular}{ | p{10cm} | p{5.8cm} | p{2.0cm} | } \hline 
\textbf{Training Set} & \textbf{Testing Set} & \textbf{ TDR (\%) @ FDR = 0.1\%} \\  \hline \hline
Bonafide vs. [Play Doh + Body Double] & Bonafide vs. [Play Doh + Body Double] & $86.11$ $\pm$ $1.95$ \\ \hline
Bonafide vs. [Play Doh + Body Double + 150 Spoof EcoFlex Patches*] & Bonafide vs. [Play Doh + Body Double] & $84.68$ $\pm$ $2.66$ \\ \hline
Bonafide vs. [Play Doh + Body Double + 150 Spoof EcoFlex Patches + 15K Synthesized EcoFlex Patches] & Bonafide vs. [Play Doh + Body Double] & $\textbf{87.08}$ $\pm$ $\textbf{1.89}$ \\ \hline \hline

Bonafide vs. [Play Doh + EcoFlex] & Bonafide vs. [Play Doh + EcoFlex] & $93.61 \pm 1.31$  \\ \hline
Bonafide vs. [Play Doh + EcoFlex + 150 Spoof Body Double Patches*] & Bonafide vs. [Play Doh + EcoFlex] & $93.59$ $\pm$ $0.77$  \\ \hline
Bonafide vs. [Play Doh + EcoFlex + 150 Spoof Body Double Patches + 15K Synthesized Body Double Patches] & Bonafide vs. [Play Doh + EcoFlex] & $\textbf{94.17}$ $\pm$ $\textbf{1.03}$ \\ \hline \hline

Bonafide vs. [EcoFlex + Body Double] & Bonafide vs. [EcoFlex + Body Double] & $98.6$ $\pm$ $0.26$ \\ \hline
Bonafide vs. [EcoFlex + Body Double + 150 Spoof Play Doh Patches*] & Bonafide vs. [EcoFlex + Body Double] & $\textbf{98.88}$ $\pm$ $\textbf{0.57}$  \\ \hline
Bonafide vs. [EcoFlex + Body Double + 150 Spoof Play Doh Patches + 15K Synthesized Play Doh Patches] & Bonafide vs. [EcoFlex + Body Double] & $98.55$ $\pm$ $0.42$ \\ \hline
\end{tabular}
}
\flushleft{\footnotesize{$^*$150 Spoof EcoFlex / Body Double / Play Doh Patches are generated from only 5 Spoof EcoFlex / Body Double / Play Doh images, respectively.}}
\end{table*}

\subsection{Training procedure}
For the Crossmatch sensor in LivDet15 \cite{mura2015livdet}, spoof training dataset is fabricated using three spoof materials, namely Ecoflex, Play Doh, and Body Double. To simulate testing our method against a novel target spoof material, we train an AdaIN spoof generator on fingerprint spoofs fabricated using the other two (known) materials to learn an effective style extractor and translator amongst spoofs in those two materials.  After this, we take $k$ fingerprints of the target material (material not used in training) which we want to use for generating a synthetic spoof dataset. $30$ patches of size $96 \times 96$ are extracted and aligned from each fingerprint image and these patches become the target material patch set for our UMT-Wrapper.

We take a large number of bonafide fingerprint patches ($\sim15000$) such that the number of patches is roughly equal to the number of patches of any one of the materials in the LivDet trainset of the CrossMatch sensor. These content fingerprint patches are fed into our UMT-Wrapper for generating synthetic spoof fingerprint patches that retain the content of the bonafide patches with the style of the target material patch set. We add these synthetic fingerprint patches along with the target material patch set to our existing training data of that classifier and test on a combined set of the LivDet training and the testing splits of that target material.

For each type of experiment, we run 5 different runs where each run has a different set of style patches and different bonafide images which are used for generating synthetic spoofs. We report the means at each training epoch in Figure 4. We report the mean TDR and its standard deviation of the last 15 epochs (from 50 to 65) in Table 1 and Table 2.

\subsection{Implementation Details}
For our experiments, we use $k = 5$. For the encoder of synthetic spoof generator, we use weights pretrained on ImageNet \cite{imagenet}. The encoder is the first few layers of a VGG-19 network.
The weights of this encoder are frozen during training of the synthetic spoof generator. For the training of our synthetic spoof generator, we use $\lambda_{c} = 1.0$ and $\lambda_{s} = 10.0$.
We use the Adam optimizer \cite{adam} and a batch size of 1. The learning rate of the generator and the discriminator are both $1e-5$. For the orientation calculation for alignment of fingerprint images, we use a window of size 64 and stride of size 32. All our experiments our performed in the PyTorch framework.

For the MobileNet V2 classifier, we also used pretrained ImageNet \cite{imagenet} weights and cut off the last fully connected layer and replace with a fully connected layer for two class outputs. We also train this with an Adam optimizer with a learning rate of $1e-4$ and a batch size of 64.
\subsection{Results}
In order to demonstrate the contribution of our proposed UMT-Wrapper for making a particular spoof detector more robust towards an unknown material $M \in$ \textit{\{Eco Flex, Body Double, Play Doh\}}, we conduct three sets of cross-material experiments for each material $M$, by adopting the leave-one-out method.

In the first set of experiment, we test on the unknown material $M$ which is the material completely left-out from training and we desire to make our spoof detector more robust to. Note that for testing on material $M$, we combine the train and the test set of that material and use this as a test set in order to increase the size of the test set with respect to the live test set. In the second set, we include $150$ spoof patches extracted from only $5$ images from training set of unknown material $M$ in the training. And in the third set, we also include 15,000 synthesized patches generated using the proposed UMT-wrapper in the training. Table~\ref{tab:crossresults} presents the cross-material results for each of the three materials. It can be observed that including the $15,000$ synthesized spoof patches in training results in an improvement in TDR @ FDR = 0.1\%.

It is important to ensure that adding the synthesized spoof patches, does not degrade the performance of spoof detector against known materials. We conduct a set of three known-material experiments similar to the cross-material experiments. Table~\ref{tab:knownresults} presents the known-material results for each of the three subsets of 2 materials. It can be observed that the known-material performance is sustained by adding synthesized spoof patches in training. This indicates the positive contributions of utilizing UMT-wrapper for improving the generalization of a spoof detector while ensuring robustness to known materials.

\section{Addition of Adversarial Loss}
\begin{figure}[t]
\centering
\hspace*{-0.6cm}
\includegraphics[scale=0.6]{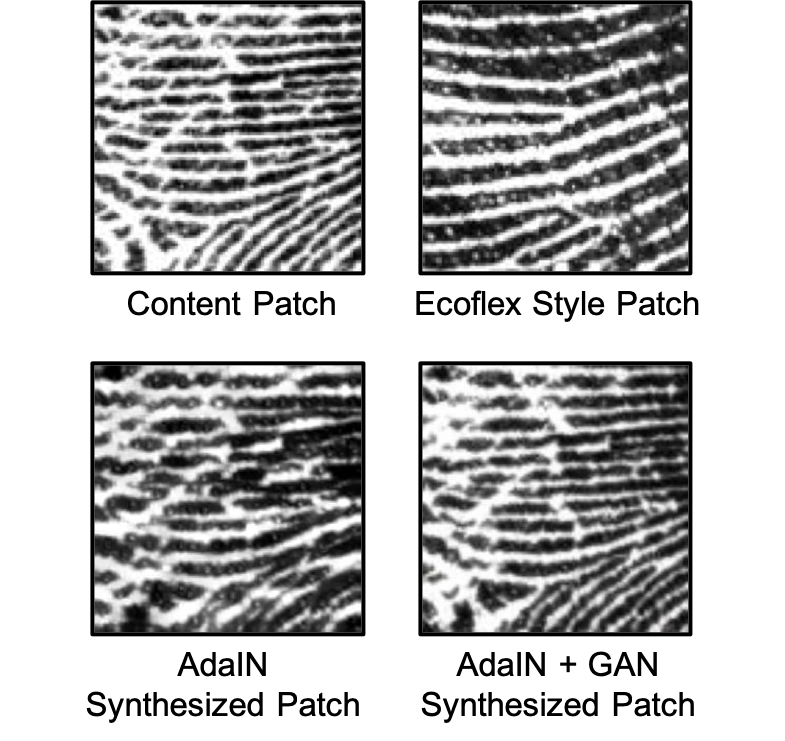}
\vspace{-0.2cm}
\caption{Effect of GAN loss on synthesized spoof patches}
\vspace{-0.2cm}
\label{fig:gan}
\end{figure}

In our work, we also experimented with the addition of an adversarial loss in our UMT-Wrapper training. This led to better visual quality of the generated spoof patches. A typical GAN setup consists of a generator $G$ and a discriminator $D$, where $D$ learns to distinguish between fake and real images and $G$ tries to fool $D$. We use a similar GAN setup to \cite{isola2016image} in our approach where bonafide patches are used to generate synthesized spoof patches similar to existing spoof patches via a discriminator. We believe that UMT-Wrapper + GAN supervision generates spoof fingerprint patches which are superior than normal UMT-Wrapper in terms of Level 1 and Level 2 features (see Figure~\ref{fig:gan}). However, there was a reduction in the TDRs obtained when adding an adversarial loss when compared to the experiments where adversarial loss wasn't added. We reason that GANs introduce certain artefacts and noise in the generated images which can be easily picked up by detectors and since we pose spoof detection as a binary problem, it turns out to be a problem to the current setup.

\section{Conclusions and Future Work}
This study proposes a framework for Universal Material Translator to generate large synthetic spoof datasets of new materials using only bonafide images and a few samples of target material type. It is shown that the proposed approach improves the generalization performance of spoof detector on unknown materials while maintaining high performance on known materials. We believe our work can provide a direction to make spoof detection systems generalize the understanding of what a spoof is. In our future work, we would like to connect our approach to adversarial machine learning and lifelong learning methods.

\section{Acknowledgement}
This research is based upon work supported in part by the Office of the Director of National Intelligence (ODNI), Intelligence Advanced Research Projects Activity (IARPA), via IARPA R\&D Contract No. 2017 - 17020200004. The views and conclusions contained herein are those of the authors and should not be interpreted as necessarily representing the official policies, either expressed or implied, of ODNI, IARPA, or the U.S. Government. The U.S. Government is authorized to reproduce and distribute reprints for governmental purposes notwithstanding any copyright annotation therein.




\small
\begin{thebibliography}{8}

\bibitem{baldisserra2006fake}
D.~Baldisserra, A.~Franco, D.~Maio, and D.~Maltoni, ``Fake fingerprint
  detection by odor analysis,'' in \emph{Proc. ICB}.\hskip 1em plus 0.5em minus
  0.4em\relax Springer, 2006
  
  \bibitem{sketch2photo}
T.~Chen, M.~ming Cheng, P.~Tan, A.~Shamir, and S.~min Hu, ``{Sketch2photo: internet image montage}'', {\em ACM SIGGRAPH Asia}, 2009.

\bibitem{chugh2018fingerprint}
T.~Chugh, K.~Cao, and A.~K. Jain, ``Fingerprint Spoof Buster: Use of Minutiae-centered Patches", in the \emph{IEEE TIFS}, 2018.

\bibitem{chugh2019fingerprint}
T. Chugh and A. K. Jain. ``{Fingerprint Presentation Attack Detection: Generalization and Efficiency}," arXiv preprint arXiv:1812.11574 (2018).

\bibitem{ding2016ensemble}
Y.~Ding and A.~Ross, ``{An ensemble of one-class SVMs for fingerprint spoof
  detection across different fabrication materials},'' in \emph{Proc. IEEE
  WIFS}, 2016
  
\bibitem{condin}
V. Dumoulin, J. Shlens, and M. Kudlur, ``{A learned representation
for artistic style}," in \emph{ICLR}, 2017

\bibitem{efros_texture}
A. A. Efros and T. K. Leung, ``Texture synthesis by non-parametric sampling,'', in \emph{ICCV}, 1999.

\bibitem{elgammal2017can}
A.~Elgammal, B.~Liu, M.~Elhoseiny, and M.~Mazzone, ``Can: Creative adversarial networks, generating ``art" by learning about styles and deviating from style norms", \emph{arXiv preprint arXiv:1706.07068}, 2017.

\bibitem{tpamijosh}
J. Engelsma, K. Cao, and A.K. Jain, ``RaspiReader: Open Source Fingerprint Reader,'' in \emph{IEEE TPAMI}, 2018

\bibitem{gatyssynth}
L.A. Gatys, A.S. Ecker, M. Bethge, ``{Texture synthesis using convolutional neural
networks}," in \emph{NIPS}, 2015.

\bibitem{neuralstyle}
L.A. Gatys, A.S. Ecker, M. Bethge, ``{A neural algorithm of artistic style}," arXiv
preprint arXiv:1508.06576, 2015.

\bibitem{ghiani2013fingerprint}
L.~Ghiani, A.~Hadid, G.~L. Marcialis, and F.~Roli, ``{Fingerprint liveness
  detection using Binarized Statistical Image Features},'' in \emph{BTAS}, 2013.

\bibitem{ghiani2012fingerprint}
L.~Ghiani, G.~L. Marcialis, and F.~Roli, ``Fingerprint liveness detection by
  local phase quantization,'' in \emph{ICPR}, 2012.

\bibitem{ghiani2017review}
L.~Ghiani, D.~A. Yambay, V.~Mura, G.~L. Marcialis, F.~Roli, and S.~A.
  Schuckers, ``{Review of the Fingerprint Liveness Detection (LivDet)
  competition series: 2009 to 2015},'' \emph{Image and Vision Computing}, 2017.

\bibitem{adainmain}
X. Huang and S. Belongie, ``Arbitrary Style Transfer in Real-time with Adaptive Instance Normalization,'', in \emph{ICCV}, 2017.

\bibitem{isola2016image}
P.~Isola, J.-Y. Zhu, T.~Zhou, and A.~A. Efros, ``{Image-to-image translation with conditional adversarial networks}," in \emph{CVPR}, 2017.

\bibitem{johnson2016perceptual}
J.~Johnson, A.~Alahi, and L.~Fei-Fei, ``{Perceptual losses for real-time style transfer and super-resolution}," in {\em ECCV}, 2016.

\bibitem{johnsonstyle}
J. Johnson, A. Alahi, and L. Fei-Fei, ``{Perceptual losses for
real-time style transfer and super-resolution}," in \emph{ECCV}, 2016.

\bibitem{adam}
D.P. Kingma, J. Ba, ``{Adam: A Method for Stochastic Optimization}," in \emph{ICLR}, 2015.

\bibitem{lapsley1998anti}
P.~D. Lapsley, J.~A. Lee, D.~F. Pare~Jr, and N.~Hoffman, ``Anti-fraud biometric
  scanner that accurately detects blood flow,'' {US Patent 5,737,439, 1998}.

\bibitem{listyle}
C. Li and M. Wand, ``{Precomputed real-time texture synthesis
with markovian generative adversarial networks}," in \emph{ECCV}, 2016.

\bibitem{demystneural}
Y. Li, N. Wang, J. Liu, X. Hou, ``{Demystifying Neural Style Transfer},'' in \emph{IJCAI}, 2017.

\bibitem{marasco2011robustness}
E.~Marasco and C.~Sansone, ``{On the Robustness of Fingerprint Liveness
  Detection Algorithms against New Materials used for Spoofing},'' in
  \emph{Proc. Intl. Conf. Bio-Insp. Syst. Sign. Process.}, 2011

\bibitem{marasco2012combining}
E.~Marasco and C.~Sansone, ``Combining perspiration-and morphology-based static
  features for fingerprint liveness detection,'' \emph{Pattrn. Reco. Letters, 2012}
  
\bibitem{marcialis2010analysis}
G.~L. Marcialis, F.~Roli, and A.~Tidu, ``Analysis of fingerprint pores for
  vitality detection,'' in \emph{Proc. 20th ICPR, 2010.}

\bibitem{mura2015livdet}
V.~Mura, L.~Ghiani, G.~L. Marcialis, F.~Roli, D.~A. Yambay, and S.~A.
  Schuckers, ``{LivDet 2015 - Fingerprint liveness detection competition
  2015},'' in \emph{Proc. IEEE 7th Intl. Conf. BTAS}, 2015

\bibitem{MinNet}
D. Nguyen, K. Cao and A.K. Jain, ``{Robust Minutiae Extractor: Integrating Deep Networks and Fingerprint Domain Knowledge}," in \emph{ICB}, 2018.
  
\bibitem{nogueira2016fingerprint}
R.~F. Nogueira, R.~de~Alencar~Lotufo, and R.~C. Machado, ``{Fingerprint
  Liveness Detection Using Convolutional Neural Networks},'' \emph{IEEE TIFS}, 2016.

\bibitem{otsu}
N. Otsu, ``{A threshold selection method from gray-level histogram}, {\em IEEE Transactions on System Man Cybernetics}, 1979.

\bibitem{pala2017deep}
F.~Pala and B.~Bhanu, ``{Deep Triplet Embedding Representations for Liveness
  Detection},'' in \emph{Deep Learning for Biometrics. Advances in Computer
  Vision and Pattern Recognition.}
  
\bibitem{rattani2015open}
A.~Rattani, W.~J. Scheirer, and A.~Ross, ``Open set fingerprint spoof detection
  across novel fabrication materials,'' \emph{IEEE TIFS}, 2015.
  
\bibitem{patchmatch}
A.~Roy, N.~Memon, and A.~Ross, ``Masterprint: Exploring the vulnerability of
  partial fingerprint-based authentication systems,'' \emph{IEEE TIFS}, 2017.

\bibitem{imagenet}
O. Russakovsky, J. Deng, H. Su, J. Krause, S. Satheesh, S. Ma, Z. Huang, A. Karpathy, A. Khosla , M. Bernstein, A.C. Berg and L. Fei-Fei, ``{ImageNet Large Scale Visual Recognition Challenge}," in \emph{IJCV}, 2015.

\bibitem{mobilenetv2}
M. Sandler, A. Howard, M. Zhu, A. Zhmoginov and L. Chen, ``{MobileNetV2: Inverted Residuals and Linear Bottlenecks}," in \emph{CVPR}, 2018.

\bibitem{incv3}
C. Szegedy, V. Vanhoucke, S. Ioffe, J. Shlens, Z. Wojna, ``{Rethinking the Inception Architecture for Computer Vision}," in \emph{CVPR}, 2016.

\bibitem{ulyanovstyle}
D. Ulyanov, V. Lebedev, A. Vedaldi, and V. Lempitsky, ``{Texture networks: Feed-forward synthesis of textures and stylized images}," in \emph{ICML}, 2016.

\bibitem{improvedtextnetworks}
D. Ulyanov, A. Vedaldi, and V. Lempitsky, ``{ Improved texture
networks: Maximizing quality and diversity in feed-forward
stylization and texture synthesis}," in \emph{CVPR}, 2017.

\bibitem{wangstyle}
X. Wang, G. Oxholm, D. Zhang, and Y.-F. Wang, ``{Multimodal transfer: A hierarchical deep convolutional neural network for fast artistic style transfer}," arXiv preprint
arXiv:1612.01895, 2016.

\bibitem{texturegan}
W. Xian, P. Sangkloy, V. Agrawal, A. Raj, J. Lu, C. Fang, F. Yu and J. Hays, ``{Texturegan: Controlling deep image synthesis with texture patches}," in \emph{CVPR}, 2017.








  





\end{thebibliography}

\end{document}